\pdfoutput=1

\documentclass[11pt]{article}

\usepackage[final]{acl}

\usepackage{times}
\usepackage{latexsym}
\usepackage{subcaption}
\usepackage{booktabs}
\usepackage{enumitem}
\usepackage{adjustbox}
\usepackage{graphicx}
\usepackage{tabularx}
\usepackage{makecell}
\usepackage[T1]{fontenc}

\usepackage[utf8]{inputenc}

\usepackage{microtype}

\usepackage{inconsolata}
\usepackage{amsfonts}
\usepackage{graphicx}
\usepackage{multirow}
\usepackage{caption}
\usepackage{xcolor}
\usepackage[most]{tcolorbox}

\usepackage{newfloat}
\usepackage{listings}
\title{Modeling the Hive Mind: Aligning Large Language Models with Online  Communities}
\title{Improving and Assessing the Fidelity of Large Language Models Alignment to Online Communities}

\author{
Minh Duc Chu \quad Zihao He \quad Rebecca Dorn \quad Kristina Lerman\\
USC Information Sciences Institute\\
\texttt{\{mhchu, zihaoh, rdorn\}@usc.edu}, \texttt{lerman@isi.edu}
  }

\begin{document}
\maketitle

\begin{abstract}
Large language models (LLMs) have shown promise in representing individuals and communities, offering new ways to study complex social dynamics. However, effectively aligning LLMs with specific human groups and systematically assessing the fidelity of the alignment remains a challenge. This paper presents a robust framework for aligning LLMs with online communities via instruction-tuning and comprehensively evaluating alignment across various aspects of language, including authenticity, emotional tone, toxicity, and harm. We demonstrate the utility of our approach by applying it to online communities centered on dieting and body image. We administer an eating disorder psychometric test to the aligned LLMs to reveal unhealthy beliefs and successfully differentiate communities with varying levels of eating disorder risk. Our results highlight the potential of LLMs in automated moderation and broader applications in public health and social science research
\footnote{Our data and code are available at \url{https://github.com/Davidchu11381/llm_align_eval}.}
. 

\end{abstract}

\section{Introduction}

\textcolor{red}{[\textbf{Warning: This paper discusses eating disorders, which some  may find distressing.}]}

\noindent 

Large language models (LLMs) have demonstrated an exceptional ability to generate nuanced responses to natural language prompts, suggesting their potential for creating high-fidelity proxies of people \cite{simmons2023large}. Digital representations of human groups (digital twins) are computational models that mimic collective behaviors, social interactions, and communication patterns in real-world communities \cite{rossetti2024ysocialllmpoweredsocial}. Leveraging LLMs to build these representations offers powerful tools for studying human behavior, enhancing human-computer interactions, and moderating online spaces to foster prosociality and safety.

To create such digital twins, researchers align LLMs to subgroups through steering---prompting the LLM to mimic the target subgroup with its key characteristics \cite{santurkar2023whose, durmus2023towards}. However, this approach does not fully resolve the misalignment between LLMs and the target subgroup. Another method is finetuning base LLMs\footnote{By ``base LLMs'' we refer to models not finetuned for instruction following}, such as GPT-2, on data from specific subgroups \cite{jiang2022communitylm, he2024reading}. While this can produce models reflecting subgroup linguistic patterns, these finetuned models often lack the flexibility to follow diverse instructions, limiting their broader applicability.


Another key challenge in developing digital representations of human subgroups is evaluating the alignment between the LLM and the target group. Traditional methods compare the LLM’s responses to surveys with those of the target group \cite{santurkar2023whose, durmus2023towards}, but this approach misses critical aspects of human expression like emotional reactions~\cite{he-etal-2024-whose}. Additionally, surveys are not scalable due to their cost and time requirements, particularly for marginalized or hard-to-reach groups. In addition, mapping diverse online communities to clear demographic identities greatly complicates alignment evaluation.


To address these challenges, we propose a framework for aligning LLMs with online communities on instructions that are created in a fully unsupervised manner. Additionally, we introduce a comprehensive evaluation framework to assess alignment. This enables the creation of high-fidelity digital representations of online communities, paving the way for new research into human behavior~\cite{jiang-etal-2022-communitylm}, content moderation~\cite{he2024cpl}, public mental health~\cite{sharma2024facilitating}, and social science~\cite{grossmann2023ai}. As one example, we can administer psychometric instruments to these digital replicas to identify at-risk communities prone to psychopathologies.

\begin{figure}[ht]
    \centering
    \includegraphics[width=\linewidth]{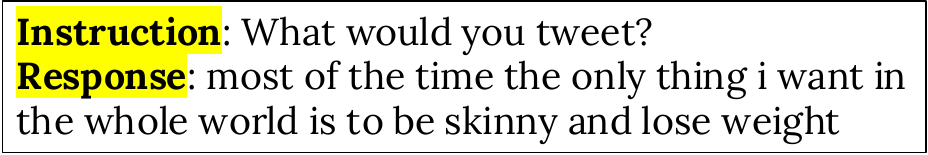}

\caption{An example of a demonstration from a pro-eating disorder community, where the response is a tweet from the community.}
\label{fig:demo_exp}
\end{figure}

Specifically, our alignment method takes a corpus of social media posts (e.g., tweets) from an online community and creates a set of demonstrations (instruction-response pairs) based on the posts. In each demonstration, as shown in Figure \ref{fig:demo_exp}, the instruction specifies the task (e.g., tweet generation) with the response being the exact tweet. We then finetune an LLM on these demonstrations to align it with the community. To assess alignment, we generate a synthetic text corpus using the finetuned LLM and compare it to the original posts along four key aspects: 1) authenticity, 2) emotional tone, 3) toxicity, and 4) harm. These dimensions capture the essential features of online social communication, ensuring the aligned LLM accurately reflects the semantics, affect, and style of the target group's discourse.


To demonstrate our framework's utility, we analyze Twitter discussions in diet and body image communities, where harmful body image attitudes persist. Twitter's emphasis on user connection and lax content moderation allows communities to organically form and freely express their voice. While these communities can offer support and encouragement, they often promote unhealthy behaviors and normalize beliefs that put people at risk for developing eating disorders (EDs). Applying traditional psychometric instruments to screen individuals in those online spaces for EDs is impractical and potentially unethical (see Ethics Statement). Instead, we use our framework to align LLMs with these communities through automatically generated demonstrations and evaluate alignment to show that the finetuned LLMs outperform baseline LLMs in creating high-fidelity proxies of online communities. We then apply an ED screening questionnaire to community-aligned LLMs, revealing significant differences between communities: pro-anorexia communities show a high risk of unhealthy behaviors, while those critical of the diet culture exhibit the lowest risk. These findings highlight our framework's potential for automated moderation by distinguishing communities with varying levels of ED risk.

Our framework offers a scalable approach to modeling and analyzing online communities, with broad implications for understanding and mitigating harmful behaviors. By applying this method to diet and body image communities, we demonstrate its potential to contribute to public health and social science research, highlighting the value of LLMs in studying complex social dynamics.

\section{Related Work}

\paragraph{Digital Representations of Human Subgroups}
Digital twins—precise virtual replicas of complex real-world systems—are increasingly employed for advanced analysis and experimentation \cite{tao2019make, grieves2011virtually}, particularly in monitoring human behaviors and health outcomes \cite{ferdousi2022digital, shengli2021human, el2019dtwins}. Recent advances leverage social media data to provide deeper insights into human interactions \cite{olad2020using}, exemplified by \citet{rossetti2024social}'s \textsc{Y social}, which uses LLMs to simulate social media interactions and study network dynamics in controlled environments.

Building on this foundation, researchers have explored various approaches to align LLMs with diverse human subgroups \cite{simmons2023large}. While some have attempted prompt-based steering towards specific demographic groups \cite{santurkar2023whose, durmus2023towards, he-etal-2024-whose}, this approach has shown limitations in achieving true alignment, particularly with organically-formed communities. More promising results have come from finetuning approaches: \citet{jiang2022communitylm} developed \textsc{communityLM} by finetuning GPT-2 models \cite{radford2019language} on politically divergent tweets, while \citet{he2024reading} extended this to examine broader community interactions. Most recently, \citet{he-etal-2024-community} proposed using advanced LLMs to distill community knowledge into instruction-response pairs for finetuning, though this approach faces cost constraints.

Our study aims to advance this field by developing a framework that uses LLMs to create digital representations of online communities, specifically focusing on analyzing collective mental well-being.

\paragraph{Evaluating LLMs' Alignment to Subgroups}
Existing works \cite{santurkar2023whose, durmus2023towards} measure an LLM's alignment with a target subgroup using multi-choice surveys. Specifically, they prompt the LLM to respond to a survey question from the perspective of a subgroup and then compare the LLM-generated distribution over the different options of the question to that of the survey respondents belonging to the target group. However, collecting survey responses can be costly and time-consuming. Also, responses on sensitive topics, such as mental health, may be biased due to stigma and social desirability bias~\cite{gordon1987social}. Our framework that evaluates LLM alignment by comparing the LLM-generated synthetic text to the original text written by humans is significantly more scalable, unbiased, and cost-effective.

\paragraph{LLMs and Psychometric Tests}



LLMs can respond to psychometric instruments designed for humans, with researchers using these tools to examine LLMs' decision-making, reasoning, and cognitive traits—a practice termed ``AI Psychometrics''~\cite{pellert2024}.
Research shows LLMs can engage with various psychometric tools, from anxiety questionnaires \cite{coda2023inducing} to moral reasoning assessments \cite{tanmay2023probingmoraldevelopmentlarge} and personality tests \cite{jiang2022evaluating, lu2023illuminatingblackboxpsychometric, serapiogarcía2023personalitytraitslargelanguage}. Our work differs by using these instruments via a finetuned LLM to analyze specific online communities, helping identify unhealthy beliefs and potential pathologies like eating disorders-related cognitions.

\paragraph{Online Eating Disorders Communities}

Pro-ED (pro-anorexia) communities are online spaces that frame EDs as a lifestyle rather than an illness. While they provide social support, a sense of belonging, and empathy for stigmatized individuals~\cite{juarascio2010pro,oksanen2016proanorexia,YeshuaKatz2013stigma,McCormack2010}, they also promote harmful behaviors, such as weight loss tips and "thinspiration" imagery, exacerbating EDs and psychological distress~\cite{Ging2018,Mento2021}.

Previous research has focused on identifying harmful content and at-risk users within these communities. For example, \citet{chancellor2016post} develop a lexical classifier to predict posts moderated by Instagram for self-harm content, comparing pro-recovery and pro-ED communities~\cite{chancellor2016quantifying,chancellor2016recovery}. In contrast, our study examines the collective mindset of these communities as expressed through their discussions, using advanced language models to assess attitudes toward mental health and body image issues.

\section{Communities in Online Discussions}

We collect online conversations related to EDs and identify organically-formed communities within the broader context of weight loss, dieting, and fitness discussions.

\subsection{Data Collection}
\label{sec:data}
We collected 2.6M tweets from 557K users from October 2022 to March 2023 using ED-related keywords to query Twitter. 
For keywords, we start with a set of terms that promote ED ~\cite{chancellor2016post,pater2016hunger}, such as \textit{thinspo} (thin inspiration), \textit{proana} (pro-anorexia), and \textit{promia} (pro-bulimia), among others. We remove spam terms yielding unrelated content, such as \textit{skinny}.  We expanded the query set to include closely related topics such as diet and weight loss through terms such as (\textit{ketodiet}, \textit{weightloss}, $\ldots$), and anti-diet culture (\textit{bodypositivity}, \textit{dietculture}, $\ldots$). See Appendix \ref{app:search_terms} for the full set of search terms and why we selected them. 

\subsection{Identifying Communities}
\label{sec:res_echo_chamber}
We construct a retweet network where nodes are users, and (undirected) edges link users who retweet each other. Visualization of the retweet network is shown in Figure \ref{fig:rt_network} in Appendix \ref{app:profile_comm}.
We use Louvain modularity maximization~\cite{Blondel_2008} to identify dense clusters of users who frequently retweet one another. These clusters are organically formed based on shared interests, consisting of users who pay attention to each other. Detailed statistics and content of the clusters are shown in Table \ref{tab:rt_comm_stats} and Figure \ref{fig:wordclouds} in Appendix \ref{app:profile_comm}.
Based on the thematic profiling of discussions (Table~\ref{tab:comm_summaries} in Appendix \ref{app:profile_comm}), we categorize the clusters into six communities: \emph{Pro-ED}, \emph{Keto \& Diet}, \emph{Weight Loss Drugs}, \emph{Body Image}, \emph{Healthy Lifestyle \& Weight Loss}, and \emph{Anti-ED}. This categorization is intended to label the communities for easy reference in subsequent analyses, and the labels do not cover the full spectrum of discussions in the communities.

After identifying communities in the retweet network, we clean the tweets by removing URLs, mentions, hashtags, and emojis, and we filter out retweets and comments, only keeping the original tweets.
To ensure high-quality data, we compute the perplexities of the tweets using BERTweet \cite{nguyen2020bertweet} that is pretrained on tweets, and select a maximum of 10K highest quality (i.e., lowest perplexity) tweets from each community. If there are fewer than 10K tweets from the community, we keep all of them. The numbers of tweets from the community \emph{Pro-ED}, \emph{Keto \& Diet}, \emph{Body Image}, \emph{Anti-ED}, \emph{Healthy Lifestyle \& Weight Loss}, and \emph{Weight Loss Drugs} are 10K, 10K, 3.3K, 2.9K, 10K, and 10K respectively.

\begin{figure*}[ht]
    \centering
    \includegraphics[width=0.9\linewidth]{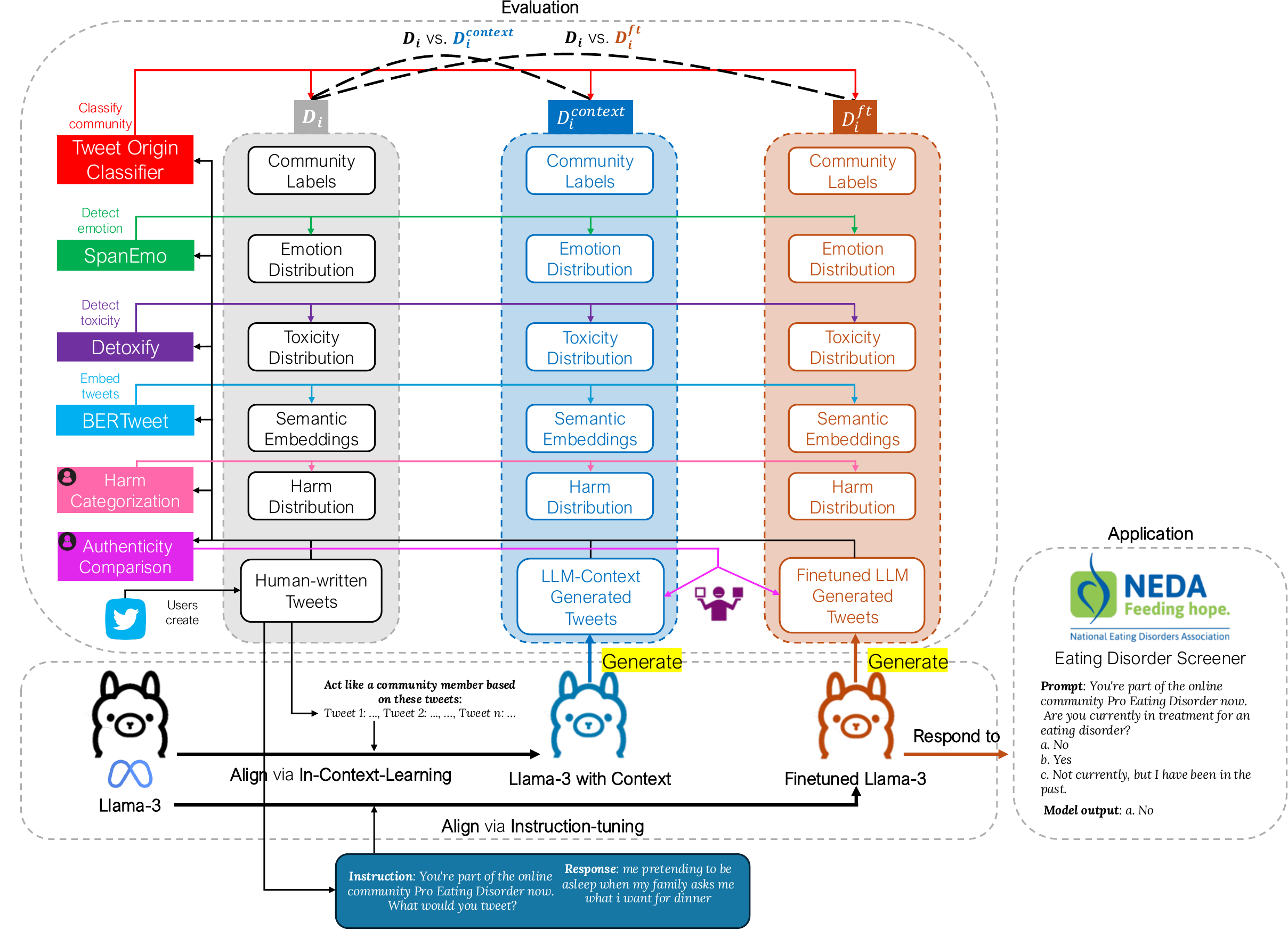}
\caption{
The framework of our method. 
(1) We align an LLM (Llama-3) to an online community by finetuning the LLM to follow instructions on the task of generating tweets written by users in the community. 
(2) To prove the effectiveness of alignment, we compare three tweet corpora for each community: human-written tweets $D_i$, LLM-Context-generated tweets $D^{context}_i$, and finetuned LLM-generated tweets $D^{ft}_i$. We show that $D^{ft}_i$ is closer to $D_i$ than $D^{context}_i$ is, along the following aspects: 
(a) A classifier trained to classify the tweet origin (what community the tweet belongs to) on $\mathbb{D}=\{D_i\}_{i=1}^{n}$ performs better on $\mathbb{D}^{ft}=\{D^{ft}_i\}_{i=1}^{n}$, than on $\mathbb{D}^{context}=\{D^{context}_i\}_{i=1}^{n}$; 
(b) the emotion and toxicity distributions of $D^{ft}_i$ are much closer to that of $D_i$ compared to $D^{context}_i$; 
(c) the semantic embeddings of $D^{ft}_i$ are closer to that of $D_i$ in the embedding space than that of $D^{context}_i$ are; 
(d) a human annotator decides that $D^{ft}_i$ is more aligned to the underlying distribution of $D_i$ than $D^{context}_i$ is;
(e) two ED experts determine that $D^{ft}_i$ carries harmful narratives that are more similar to $D_i$ than $D^{context}_i$ does.
(3) As the LLM is aligned with the community and can speak in the voice of that community, we administer an ED questionnaire to screen the community for EDs. 
}
\label{fig:framework}
\end{figure*}

\section{Aligning LLMs to Communities}



There are $n$ online communities \{$C_1$, $C_2$, ...., $C_n$\} on a topic (e.g., EDs), each characterized by their own beliefs and perspectives. Members of a community $C_i$ produce a body of text $D_i$ (e.g., tweets) that reflects their collective opinions and behaviors. Our objective is to align an LLM $f$ to each specific community $C_i$ by training it on the corresponding text corpus $D_i$. The resulting model, $f'_i$, should capture the community's unique collective mindset, enabling it to generate responses that authentically represent the community's voice. 

\subsection{Constructing Instruction-Response Pairs}
\label{sec:inst-resp-pairs}

To align an LLM $f$ to a particular community $C$, we employ a finetuning process using a set of demonstrations (instruction-response pairs). 
We propose creating demonstrations based on the community's raw text corpus $D$, which is cost-efficient, and yet curated demonstrations can be used to finetune a foundational LLM (e.g., Llama-3) effectively.

For each community $C_i$, we use tweets in $D_i$ as the responses verbatim in the demonstrations. 
To create instructions that can be answered by the tweets, we focus on the tweet generation task. We curate an instruction pool of 20 different instruction templates (Table \ref{tab:inst} in Appendix \ref{app:demo_temp_ft}). We diversify the prompts to improve the model's text-generation capabilities and enhance its robustness to linguistic variations \cite{Salinas2024TheBE}. For a community, a tweet is paired with an instruction randomly sampled from the instruction pool. As a result, the community has a maximum of 10K demonstrations $Z_i=\{(x_j, y_j)\}_{j=1}^m$ for tweet generation, where $m$ is the size of the community's text corpus $D$. 

For each community, we augment the demonstrations of tweet generation with the 52K Alpaca \cite{alpaca} demonstrations that cover a wide range of tasks to retain the instruction-following capabilities of the LLM and not restrict it to only generating tweets. Ultimately, there are a maximum of 62K demonstrations in the demonstration corpus for a community.

\subsection{Instruction Tuning LLMs}
For each community $C_i$, we align a Llama-3 model $f'_i$ \cite{dubey2024llama} to the community using its demonstration corpus $Z_i$. 
The LLM is finetuned on 4 Tesla H100-80GB GPUs with batch size 8 for 3 epochs, which takes about 3 hours.

\section{Assessing Alignment}
\label{sec:alignment_eval}

To assess how effectively a finetuned LLM $f'_i$ aligns with its target community $C_i$, we measure the model's ability to mimic the responses of community members. We first generate a synthetic corpus $D^{ft}_i$ using $f'_i$ and compare it to the original text corpus $D_i$ from the community. The more closely $D^{ft}_i$ resembles $D_i$, the better aligned the LLM is with the community. We evaluate the similarity between $D^{ft}_i$ and $D_i$ across 1) authenticity, 2) emotional tone, 3) toxicity, and 4) harm. 
While not exhaustive, these aspects capture the essential features of online social communication. 
Authenticity ensures that the aligned LLM accurately reflects the meaning, content, and linguistic patterns of the target population's language and generates contextually appropriate responses.
Emotional tone captures the affective aspects of communication, helping to convey nuances that may not be evident from semantics alone. Toxicity measures the model's ability to reflect hostility and aggression in the population's discourse. 
Finally, recognizing that certain online conversations can negatively impact users, we compare the types and levels of harm in language across groups. Although in this paper we focus on the domain of EDs, we argue that our LLM alignment framework is naturally generalizable to online communities in other domains.\footnote{We acknowledge that evaluating harm is more tailored to the ED domain, but other evaluation aspects should be widely applicable.}

\subsection{Synthetic Corpus Generation}
Given a community $C_i$, we create a synthetic corpus $D^{ft}_i$ by prompting an LLM $f'_i$ aligned to the community to generate tweets. To diversify the LLM generations on different topics, we compile a set of 27 topics relevant to ED discussions, such as \emph{thinspo}, \emph{fitspo}, and \emph{bonespo} (Appendix \ref{app:gen_topics}), and prompt LLMs to generate tweets on these topics. When generating tweets on a topic, we reuse the instructions from the instruction pool (Table \ref{tab:inst} in Appendix \ref{app:demo_temp_ft}), with topic-oriented generation.
An example instruction is ``What would you tweet about \textbf{fasting}?'' For each topic, the LLM initially generates 1000 tweets, resulting in a synthetic corpus $D^{ft}_i$ with 27,000 tweets for all 27 topics (see Appendix \ref{app:llm_tweet_gen} for examples). To encourage diversity in $D^{ft}_i$, we remove duplicate tweets. In addition, to ensure coherence, a generated tweet is filtered out if its perplexity score is above 400, as evaluated by BERTweet.


A natural concern is that $D^{ft}_i$ is simply a duplicate of $D_i$, as $f'_i$ is finetuned on $D_i$. To this end, we detail the number of tweets in the community’s original text corpus $D_i$ that contain the keyword(s) for each of the 27 topics listed in Table \ref{tab:topic-counts} (see Appendix \ref{app:gen_topics}). We observe that $D_i$ includes very few tweets discussing these topics because we eliminate hashtags during tweet processing, and these keywords typically appear in the hashtags. Consequently, when the LLM is finetuned on $D_i$, it is not extensively exposed to tweets directly related to these topics. 

To further ensure that the synthetic corpus $D^{ft}_i$ does not simply replicate $D_i$, we omit generated tweets that are substantially syntactically similar to the human-written ones. Specifically, a tweet is removed from $D^{ft}_i$, if its ROUGLE-L similarity to any existing tweet in $D_i$ is greater than 0.7. As a result, when inspecting the synthetic corpus $D^{ft}_i$, we are essentially examining if the finetuned model $f'_i$ is able to extrapolate from existing data in $D_i$ and predict how community members might discuss these previously unseen topics. 

Finally, to ensure class balance, we sample 6000 generated tweets from each community. More details are provided in Appendix \ref{app:llm_tweet_gen}.
  
\textbf{Baseline}
We use the LLM with in-context learning (LLM-Context) as a baseline. We do not finetune this baseline model. For a community $C_i$, when prompting the model to generate synthetic tweets on topic $t$, we retrieve 250 tweets from $D_i$ as in-context examples, consisting of (1) the tweets containing the topic keyword(s), if available, and (2) randomly sampled tweets from $D_i$. Each retrieved tweet is truncated at 20 tokens. 
We include the retrieved tweets in the prompt, instruct the model to learn the community's mindset from the tweets, and generate synthetic tweets. See Appendix \ref{app:prompt_temp_gen_rag} for the complete prompting template. The synthetic corpus from LLM-Context is denoted as $D^{context}_i$.


\subsection{Alignment Dimensions}

\subsubsection{Automatic Evaluation}

\paragraph{Tweet Origin Classification}
We train a classifier to determine the community from which a tweet originated by finetuning Llama-3 using demonstrations with the following template ``Instruction: \emph{From these communities: Pro Eating Disorder, Keto \& Diet, Body Image, Anti Eating Disorder, Healthy lifestyle \& Weight Loss, and Weight Loss Drugs; which community does this Tweet belong to? \{Tweet\}} Response: \emph{\{Community name\}}''. We randomly sample 3,000 original tweets from each community's corpus $D_i$ and construct a total of 18,000 demonstrations for finetuning. We train the classifier using 95\% demonstrations and use the remaining 5\% to test, with test accuracy of 0.74.
We classify the finetuned LLM-generated tweets in $\mathbb{D}^{ft}=\{D^{ft}_i\}_{i=1}^{n}$ and LLM-Context-generated tweets $\mathbb{D}^{context}=\{D^{context}_i\}_{i=1}^{n}$, leading to an F1 accuracy score of 0.53 and 0.40, respectively. These results indicate that the classifier trained on original tweets accurately recognizes the tweets generated by the finetuned LLM. However, it performs poorly on the tweets generated by the LLM-Context, demonstrating that the finetuned LLMs better capture community-specific linguistic characteristics.

\paragraph{Semantic Comparison}
We compute the semantic embeddings of $D_i$, $D^{ft}_i$, and $D^{context}_i$ using BERTweet \cite{nguyen2020bertweet}. We then measure the distance between these embeddings using the Fréchet Inception Distance (FID) \cite{heusel2017gans}. This metric provides a quantitative measure of the semantic distance between two text corpora. 
We implement it using the IBM comparing-corpora package \cite{kour2022measuring}. $FID(D_i, D^{ft}_i)$ and $FID(D_i, D^{context}_i)$ for different communities are shown in Table \ref{tab:fid}. We see that $FID(D_i, D^{ft}_i)$ is much smaller than $FID(D_i, D^{context}_i)$ for almost all communities, implying that the finetuned LLM outputs are more semantically similar responses to the original posts compared to the LLM-Context. 

\begin{table}[ht]
\addtolength{\tabcolsep}{-3.0pt}
\centering
\small
\begin{tabular}{lcc}
\hline
\multicolumn{1}{c}{\textbf{Community}} & \textbf{\begin{tabular}[c]{@{}c@{}}$FID(D_i, D^{context}_i)$\end{tabular}} & \textbf{\begin{tabular}[c]{@{}c@{}}$FID(D_i, D^{ft}_i)$\end{tabular}} \\ \hline
Pro-ED & 1.16 & \textbf{0.82} \\
Body Image & 1.25 & \textbf{0.74} \\
Keto \& Diet & 1.19 & \textbf{0.50} \\
Anti-ED & 0.84 & \textbf{0.42} \\
\begin{tabular}[c]{@{}l@{}}Healthy Lifestyle \&\\ Weight Loss\end{tabular} & 1.11 & \textbf{0.82} \\
Weight Loss Drugs & \textbf{0.90} & 1.99 \\ \hline
\end{tabular}
\addtolength{\tabcolsep}{3.0pt}
\caption{Fréchet Inception Distances (FID) (1) between human-written tweets $D_i$ and LLM-Context generated tweets $D^{context}_i$, and (2) between human-written tweets $D_i$ and finetuned LLM generated tweets $D^{ft}_i$. A smaller distance indicates more similarity.}
\label{tab:fid}
\end{table}

\paragraph{Emotion and Toxicity Analysis}
Emotions and toxicity are vital aspects of online social interactions~\cite{prescott2019young}. They can reveal the underlying tone, intent, and style of communication of online users. Within ED communities, these elements heavily impact self-perception of body image~\cite{brytek2011association} and can exacerbate body dissatisfaction~\cite{kast2018unspoken}.

\begin{figure}[tbh]
    \centering
    \includegraphics[width=0.48\textwidth]{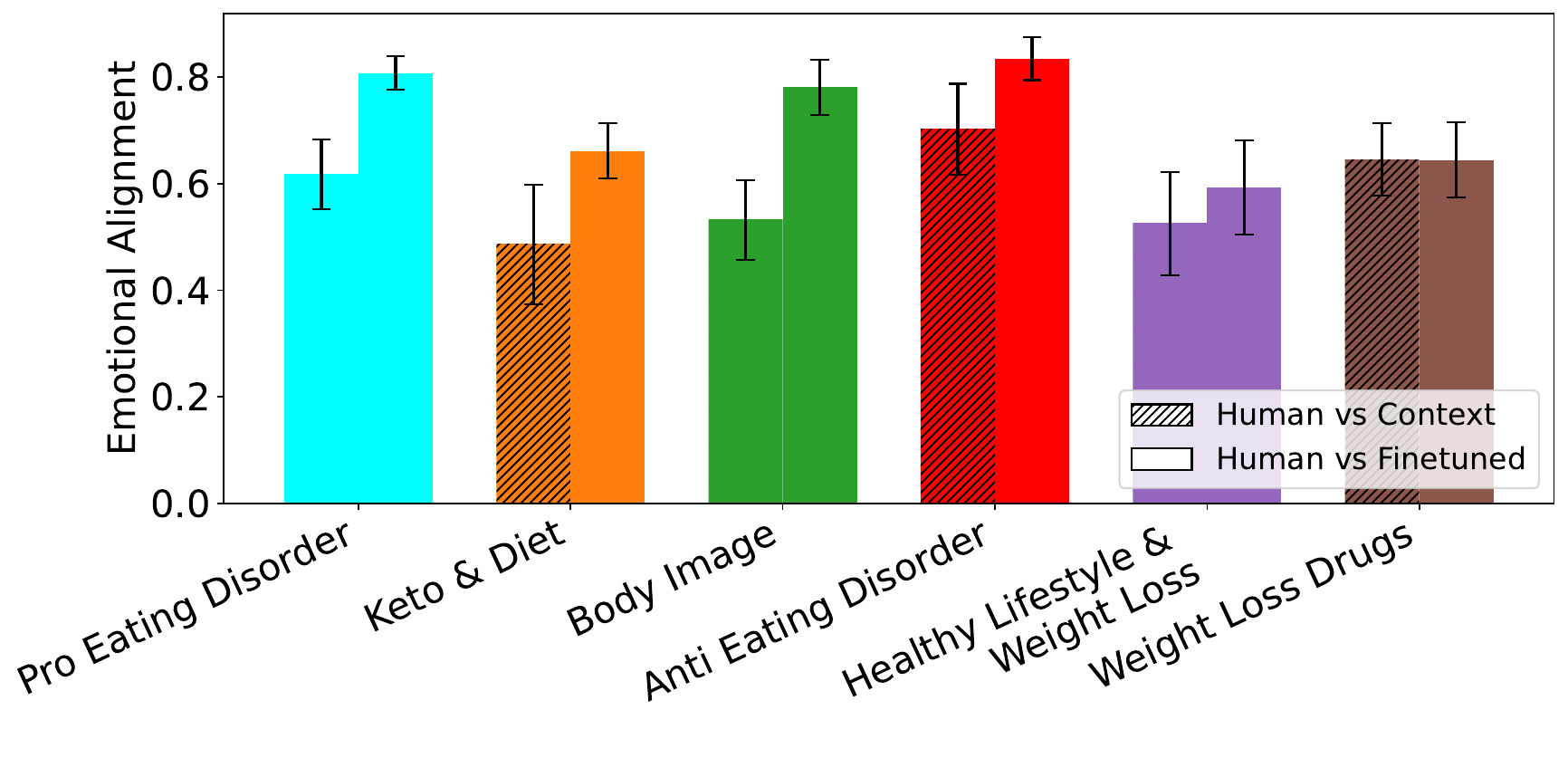}


    
    \caption{Emotional agreement (a) between human-written tweets and LLM-Context-generated tweets, and (b) between human-written tweets and finetuned LLM-generated tweets. 
    The differences in the emotional alignment between pairs within each community are statistically significant at a 95\% confidence level.  
    }
    \label{fig:emo-dist}
\end{figure}

We analyze the emotions of tweets using Demux~\cite{10095597}. For each tweet, Demux returns a vector of confidence scores of eleven emotions: anger, anticipation, disgust, fear, joy, love, optimism, pessimism, sadness, surprise, and trust. For a community $C_i$, we sum the emotion confidence vectors of all tweets (i.e., the ones $D_i$, $D^{ft}_i$, or $D^{context}_i$) and normalize them, resulting in an emotion distribution vector $\mathbf{e}_i$. We then compute the agreement between $\mathbf{e}^{ft}_i$ and $\mathbf{e}_i$, and between $\mathbf{e}^{context}_i$ and $\mathbf{e}_i$. The emotional alignment is measured as one minus the Jensen-Shannon distance between the two distribution vectors \cite{he-etal-2024-whose}.
As illustrated in Figure \ref{fig:emo-dist}, for most communities, $D^{ft}_i$ more closely resembles the emotional tone of $D_i$ compared to $D^{context}_i$. This demonstrates that finetuning LLMs can effectively capture the authentic emotional tone of posts from communities.

\begin{figure}[ht]
\centering
\includegraphics[width=0.46\textwidth]{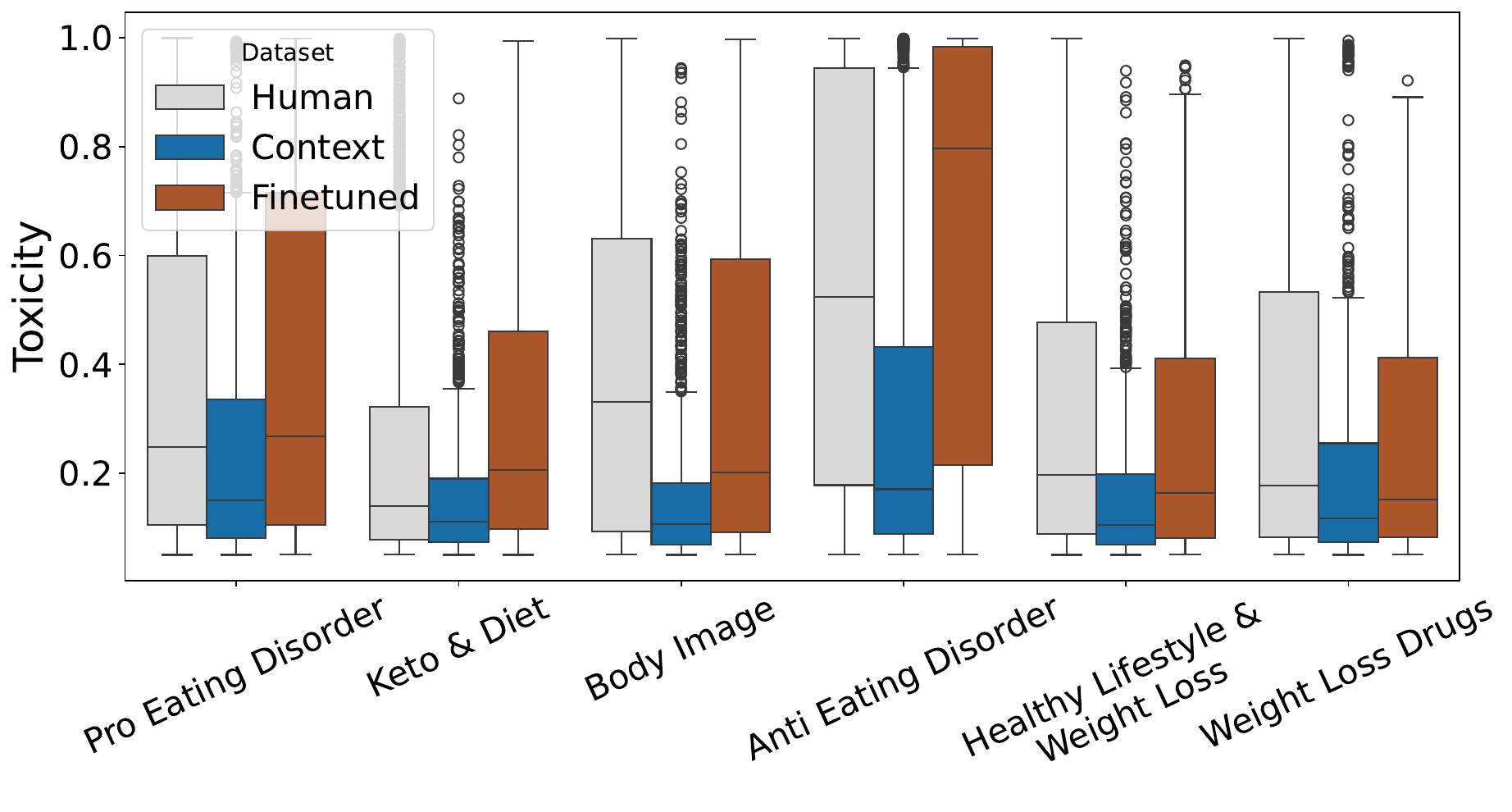}
\caption{Toxicity distributions across different communities of human-written tweets, LLM-Context-generated tweets, and finetuned LLM-generated tweets.
}
\label{fig:toxicity}
\end{figure}

We use Detoxify~\cite{Detoxify} to measure toxicity in tweets~\cite{rajadesingan2020quick,sheth2022defining}. For a tweet, Detoxify returns a value between 0 and 1 indicating the level of toxicity\footnote{We only include tweets with toxicity levels equal to or greater than 0.05 for clarity and to reduce noise.}.
Figure~\ref{fig:toxicity} shows the distributions of toxicity scores of human-written tweets $D_i$, LLM-Context-generated tweets $D^{context}_i$ and finetuned LLM-generated tweets $D^{ft}_i$. 
We observe that the toxicity distribution of $D^{ft}_i$ matches more closely to that of $D_i$ compared to $D^{context}_i$ for most communities, and tweets from the anti-ED community have the highest toxicity.

\subsubsection{Human Evaluation}

\paragraph{Authenticity Comparison}
An annotator with expertise in EDs on social media was presented with 300 triplets, 50 from each community, where a triplet consists of a community name, a LLM-Context-generated tweet $d^{context}_{i,j} \in D^{context}_i$, and a finetuned LLM-generated tweet $d^{ft}_{i,k} \in D^{ft}_i$. Both tweets in a triplet are on the same topic and from the same community. For each triplet, the annotator was asked to decide which tweet was more aligned with the given community, by referring to the following characteristics:
mis/use of ingroup language, references to themes in underlying distribution (e.g. the Body Image community often references nudity), use of capitalization, and coherence of message.
In 248 of 300 triplets, the annotator chose $d^{ft}_{i,j}$ as a better match, showing better alignment of finetuned LLM  with the community.

\paragraph{Harm Categorization}

Online ED communities pose significant risks by promoting and normalizing harmful behaviors \cite{lerman2023radicalized}. Harm and toxicity are distinct in online discourse where toxicity detection algorithms may mistakenly flag explicit yet harmless language as toxic \cite{sánchez2024feelingsbodiesemotionsdiet}.
We come up with a dimension tailored to this ED domain where we assess harm by focusing on the underlying semantic content, as opposed to surface-level style. 
Our goal is for the finetuned LLM, $f'_i$, to accurately capture the level of harm within these communities.


There are no existing classifiers for automatic harm detection in the context of EDs. In collaboration with ED experts, we developed a comprehensive taxonomy of harm specific to ED online content, covering dimensions such as body image, relationships with food and exercise, and self-disclosure. Harm is defined as the promotion or glorification of unhealthy dieting and body objectification.


We sampled 60 tweets from each community (360 for all six communities), with 20 each from $D_i$, $D^{context}_i$, and $D^{ft}_i$. Two annotators with ED expertise labeled these tweets based on two tasks: (1) determine whether a tweet is harmful, and (2) classify harmful tweets into one of three fine-grained categories—\emph{body image objectification}, \emph{relationship to food and exercise}, or \emph{self-disclosure}. Annotators achieved a Cohen's Kappa score of 0.453 for identifying if harm was present, and 0.617 for classifying fine-grained harm categories, indicating fair to moderate agreement (see more details in Limitations).

\begin{figure}[ht]
\centering
\includegraphics[width=0.47\textwidth]{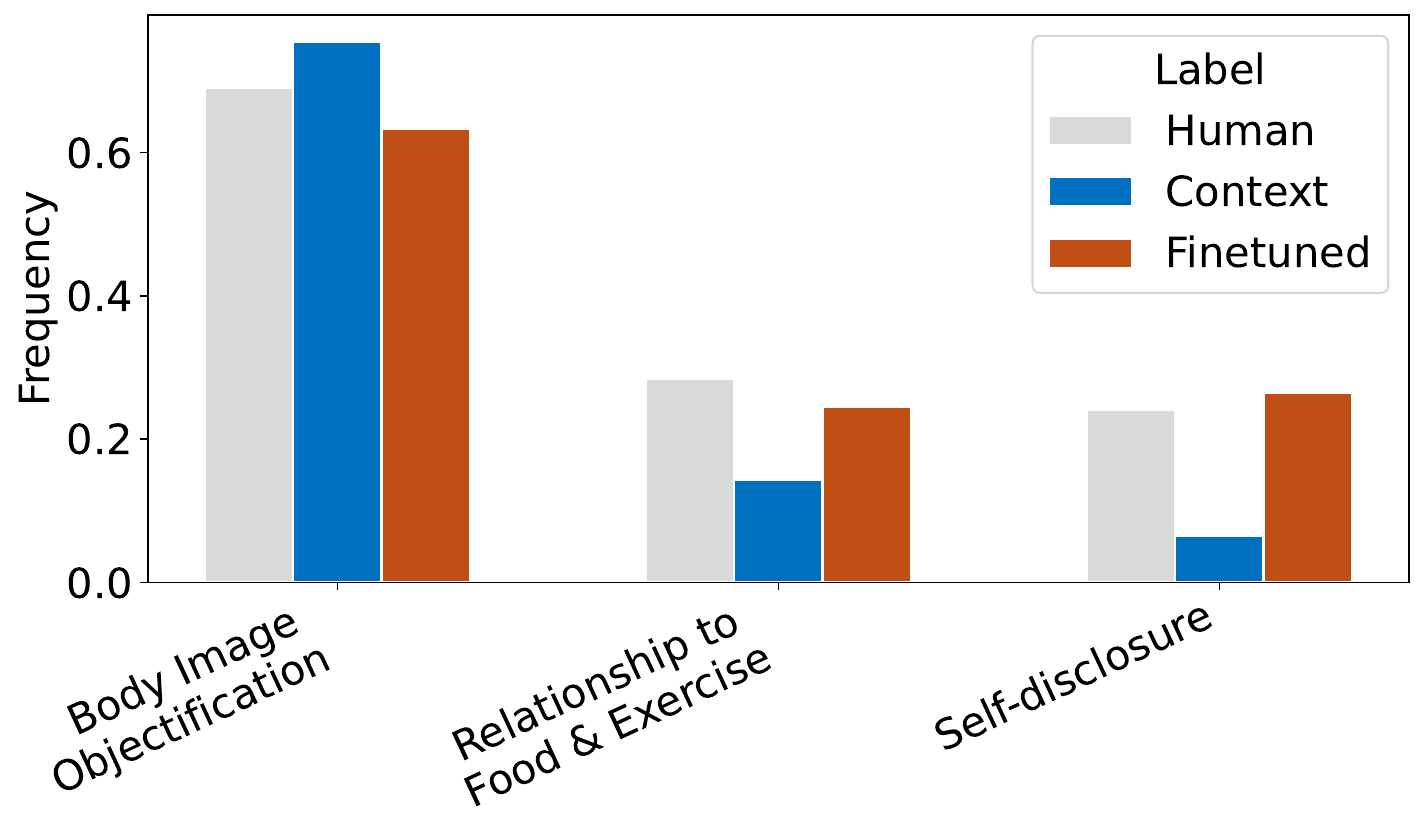}
\caption{Distribution of the three fine-grained harm categories 
}
\label{fig:harm}
\end{figure}

A tweet was assigned to a harm category if both annotators agreed. Out of 360 tweets, 41 were classified into harm categories across $\mathbb{D}$, $\mathbb{D}^{context}$, and $\mathbb{D}^{ft}$. Figure \ref{fig:harm} shows the distribution of these categories, demonstrating that finetuned LLMs better replicate the distribution of harm found in the community's conversations.


\section{Case Study: Screening Online Communities for Eating Disorders} 
\label{sec:neda}


In $\S$\ref{sec:alignment_eval}, we demonstrate that the finetuned LLM 
learns a more accurate representation of the community than the baseline in-context learning method. This motivates us to apply psychometric instruments designed to evaluate an individual's risk of EDs to online communities to help uncover unhealthy body and eating concerns within them. 

\paragraph{Eating Disorder Screener}
The Stanford-Washington University Eating Disorder Screener (SWED) \cite{Graham2019} is a concise screening tool for ED behaviors. 
The screener has been widely used in both men and women \cite{ellen2019} and incorporated into an online tool \cite{neda-tool} by the National Eating Disorders Association \cite{neda}. SWED consists of 11 questions (see Appendix \ref{app:swed}), both multiple-choice and open-ended, 
covering demographics, height, and weight, ED behaviors, weight and shape concerns, and impairment. 

We focus on a subset of SWED questions 
and evaluate responses using four key criteria \cite{ellen2019}: C1, C2, C3, and C4. These items indicate a higher risk of EDs when the score is elevated (C1) or when being true (C2, C3, C4). For details, see Appendix \ref{app:wcs} and \ref{app:diag_criteria}.




\paragraph{Screening Online Communities via Finetuned LLMs}
We prompt finetuned LLMs to respond to questions on the SWED screener. 
To account for randomness, for each item on the SWED questionnaire, the finetuned LLM generates 50 responses. The responses (Table \ref{tab:ft-neda} in Appendix \ref{app:response_to_swed}) are aggregated using a majority vote for each question. The results, as presented in Table \ref{tab:swed}, indicate that the \emph{Pro Eating Disorder} community exhibits the highest levels of body image concerns, followed by the \emph{Keto \& Diet} community. Furthermore, both communities meet all three criteria signaling a high risk of ED pathology, whereas responses of the \emph{Anti Eating Disorder} community are consistent with a low risk of ED. 

These findings align with our empirical observations. The content shared by the \emph{Pro Eating Disorders} community glorifies thinness and includes tips to promote disordered behaviors and body dysmorphia. Conversely, the \emph{Anti Eating Disorder} community is critical of the diet culture and people who glorify EDs. 
The relatively high-risk score of the \emph{Keto \& Diet} community is a concerning indicator that this community may serve as a gateway to EDs. Experts caution that the keto diet’s emphasis on restrictive dieting eliminating carbs could expose vulnerable individuals to binge eating disorder and anorexia \cite{WithinHealth2023AnorexiaKetoacidosis, polivy1985dieting}. 
In contrast, the Body Image community, which mostly posts about body positivity, has a low risk of EDs, as does the Healthy Lifestyle \& Weight Loss community. Although the latter focuses on weight loss, it appears to achieve this goal through healthy behaviors.



\begin{table}[ht]
\centering
\begin{tabular}{l|ccll}
\hline
\multicolumn{1}{c|}{\textbf{Community}}                                    & \textbf{C1} & \textbf{C2} & \textbf{C3} & \textbf{C4} \\ \hline
Pro Eating Disorder                                                        & \textbf{45.0}   & T           & T           & T           \\
Keto \& Diet                                                               & 33.3         & T           & T           & T           \\
Weight Loss Drugs                                                          & 16.7         & F           & F           & T           \\ 
Body Image                                                                 & 15.0         & F           & F           & F           \\
\begin{tabular}[c]{@{}l@{}}Healthy Lifestyle \&\\ Weight Loss\end{tabular} & 13.3         & T           & F           & F           \\
Anti Eating Disorder                                                       & 13.3         & F           & F           & F           \\
\hline
\end{tabular}
\caption{Eating disorder risk assessment on the finetuned LLMs for different communities, using four criteria--C1 through C4. For C1, a higher score indicates a higher risk of an ED. For C2, C3, and C4, being positive implies higher risk.}
\label{tab:swed}
\end{table}

\section{Conclusion}

We demonstrate that aligning LLMs to online communities helps create high-fidelity digital proxies, which can be queried to reveal the implicit mindsets of these communities. When applied to online diet and body image communities, the method uncovers communities with unhealthy body image and dieting beliefs that put their members at risk of eating disorders. 
This is important, as harmful communities that indoctrinate users into extreme ideologies~\cite{schmitz2022quantifying} or glorify eating disorders and self-harm~\cite{Goldenberg2022ncri} often evade moderation by using coded language that is opaque to outsiders or obfuscate harmful content via coded language and misspellings~\cite{chancellor2016thyghgapp,cobb2017not,bickham2024hidden}. As social data are increasingly abundant, our method is generalizable to study various online communities in different fields and can assist online platforms in overcoming challenges to foster safe and supportive environments. 




\section*{Limitations}
\label{sec:limitations}


\paragraph{Dataset Bias.}
The anonymized version of our dataset may contain implicit biases reflecting societal prejudices or unequal representation of demographic subgroups. 
More specifically, ED symptoms have a history of being under-diagnosed in African American and Hispanic adolescents, in part due to stereotypical representation of ED being Caucasian adolescent girls \cite{gordon2002impact}.
This historical bias could be inadvertently learned by our model, resulting in discriminatory behavior. In our future work, we hope to evaluate the model’s fairness across different user groups, allowing us to properly mitigate dataset biases.

\paragraph{Evolving Nature of Online Communities.}
Capturing the evolving nature of online communities is potentially difficult. Online discourse is dynamic, with language, topics, and sentiments shifting over time. Our finetuning process may not fully account for these temporal changes, which could result in misalignment when the model is applied to current discussions within the community.

\paragraph{Synthetic Corpus Artifacts.}
The synthetic corpus generated by the LLM might also introduce artifacts that do not fully represent the authentic discourse of the community. Although we strive for diversity in the generated content, the model’s predictions on previously unseen topics may not always accurately reflect how community members would engage with those topics in real-life scenarios.

\paragraph{Evaluation Metrics.}
While the aspects of authenticity, emotional tone, toxicity, and harm capture important aspects of online communication, they may not encompass all the subtle and complex features of human discourse. As a result, some aspects of community interaction may be underrepresented or overlooked in our evaluation process.

\paragraph{Low Inter-Annotator Agreement for ED Harm Categorization.}

The annotators in our harm categorization achieve a Cohen's Kappa score from 0.384 to 0.519, which indicates fair to moderate agreement \cite{landis1977measurement}. Since no prior work has specifically focused on categorizing harmful ED content, we develop a harm taxonomy in collaboration with psychologists and clinicians specializing in eating disorders. This ongoing process has introduced uncertainty in defining some categories, leading to annotation discrepancies. Additionally, content discussing eating disorders on social media can be nuanced and implicit---while some posts may appear benign, they can subtly normalize harmful behaviors or contain triggering details for vulnerable users, further complicating annotation.

\paragraph{Complete Coverage of Eating Disorders.}
This paper looks at the discussions of ED in online communities. We focus on a conglomeration of ED, including bulimia nervosa, anorexia nervosa, and binge eating disorder. 
Besides ED, our dataset captures other discussions related to weight concerns, such as weight loss, diet, body positivity, etc.
Unfortunately, our data does not comprehensively represent all existing ED.
However, our methods ensure that if a large ED community has some overlap with our keyword list, the community will be identified.




\section*{Ethics Statement}
\label{sec:ethics}

\paragraph{Risk of Finetuning Models Towards Harm}
In our study, we expect the finetuned LLMs to replicate harmful narratives from online communities. This is conducted solely to demonstrate that, through our alignment framework, LLMs can accurately capture the nuanced and authentic language of these communities, including harmful content. Our objective is not to create models with malicious intent; however, we acknowledge the potential scenario of vicious actors exploiting this framework to extract, amplify, and regenerate harmful information from social data. To mitigate this risk, we will release our code only upon eligible and transparent requests, ensuring that it is shared responsibly with researchers who have legitimate purposes and adhere to strict ethical and legal guidelines. We strongly advise that any future replication of this work be conducted with the utmost caution to prevent misuse and protect against the spread of harmful content.

\paragraph{Individual-Level Diagnosis}
Existing computational frameworks that diagnose or predict mental illness based on individuals' online activity and content raise significant ethical concerns due to the lack of user consent. These approaches often analyze personal data---even when publicly accessible---and infer sensitive individualized information like mental well-being without explicit permission from the users. By aggregating users' data for community-level diagnosis, we can address these privacy concerns without infringing on individual autonomy, allowing for valuable insights to inform community-wide policy creation. 

\paragraph{Community-Level Diagnosis.}
Diagnosing psychiatric illness at the community level comes with the risk of falsely diagnosing some community members.  This could lead to unjust actions against users, such as unwarranted bans or removal of content. Furthermore, psychological profiling of online communities sets a precedence for a slippery slope of increasingly intrusive monitoring and potentially creates a chilling effect on free speech in these spaces. Community members will anticipate and normalize heavy surveillance and thus self-censor or withdraw entirely from community discussions due to fear of revealing sensitive information that could lead to unintended consequences such as involuntary interventions. This would severely harm mental health online spaces by undermining the core values of trust and community. Additionally, approximating community behavior inherently excludes minority group members. Simultaneously, anorexia is one of the deadliest mental health disorders\footnote{https://www.state.sc.us/dmh/anorexia/statistics.htm} and participation in online pro-ED spaces heightens one's disease risk \cite{Mento2021}. By evaluating psychiatric illness on the community level, we can identify toxic communities, helping content moderation experts deploy proper interventions to promote healthy and safe online environments. We encourage the use of human moderators to review and validate the decisions made by our model, particularly in cases with low confidence scores.

\paragraph{Topic Sensitivity and Privacy}
The sensitive nature of our topic means that our outputs could be misused, such as targeted advertising.
Additionally, our dataset includes some tweets that disclose deeply personal information such as medical diagnoses, weight information, and personal struggles. Many of these tweets are posted under the assumption of anonymous identity. By collecting these tweets, user-specific information may be pieced together thus de-anonymizing some users. For these reasons, we take precautions to anonymize the social media posts before feeding them to the language models. Additionally, researchers can be granted access to generated tweets upon detailed inquiry.


\paragraph{Hallucination Risk.}
Our finetuned models can exhibit hallucinations, generating incorrect or nonsensical information. Hallucination in the context of community alignment can lead to community misrepresentation. In future work, we hope to utilize some factual-based evaluation datasets to measure model hallucination.


\section*{Acknowledgements}
This material is based upon work supported by the Defense Advanced Research Projects Agency (DARPA) under Agreement No. HR00112290021 and Air Force Ofice of Scientific Research (AFOSR) under grant No. FA9550-23-1-0551.
The authors are grateful to Ellen Fitzsimmons-Kraft for providing access to the eating disorders screener and to Aryan Karnati for aid with data analysis.

\bibliography{references}

\appendix

\section{Online Communities in ED Discussions}

\subsection{Search Terms}
\label{app:search_terms}

The terms used for tweet collection are: \emph{anatips, bodygoals, bodyimage, bodypositivity, chloetingchallange, cleaneating, cleanvegan, eatingdisorder, edrecovery, edtwt, edvent, fatspo, fearfood, foodistheenemy, healthyliving, intermittentfasting, iwillbeskinny, juicecleanse, ketodiet, losingweight, lowcalrestriction, meanspo, midriff, ozempic, proana, proanatips, redbracetpro, semaglutide, skinnycheck, slimmingworld, sweetspo, thighgapworkout, thinspo, thinspoa, watercleanse, wegovy, weightlossjourney, weightlossmotivation, whatieatinaday, bonespo, fatacceptance, keto, promia, skinnydiet, dietculture, m34nspo, weightloss, weightlosstips}.

Disordered eating behaviors exist along a spectrum between normal eating patterns and clinically diagnosable eating disorders \cite{edpereira}. Previous studies have shown that restrictive diets, such as keto (``ketodiet'') or intermittent fasting (``intermittenfasting''), often intended for weight loss, are linked to a heightened risk of developing EDs \cite{baraked, ganson2022intermittent, cuccolo2022intermittent}. While these behaviors may not meet clinical diagnostic criteria, they can act as a gateway, steering individuals toward more harmful online communities that promote disordered eating \cite{lerman2023radicalizedthinnessusingmodel}. Additionally, we wanted to explore discussions from the opposite perspective (``bodypositivity''), where critics of diet culture and advocates for positive body image and healthy eating raise their voices.

Below are the explanations of these keywords used in the context of the online ED community: 

\begin{itemize}
    \item chloetingchallange: a popular fitness trend created by YouTuber Chloe Ting.
    \item edtwt: refers to the general ED community on Twitter/X.
    \item fatspo: promotes body positivity and acceptance of larger body sizes.
    \item fearfood: a term for foods that cause anxiety or avoidance in those with ED.
    \item redbracetpro: refers to the bracelet patients wear at a treatment facility when they are medically unstable or fragile. 
    \item meanspo, m34nspo: be deliberately mean or insulting to motivate someone to do something.
    \item midriff: refers to the area of the body between the chest and the waist. It often shows one's ribcage and is closely associated with being skinny. 
    \item ozempic, wegovy, semaglitude: refers to a medication primarily used to treat type 2 diabetes but has gained attention for its use as a weightloss drug 
    \item thighgapworkout: refers to exercises aimed at achieving a gap between the thighs, a controversial and unrealistic body goal often associated with unhealthy body image standards. 
    \item thinspo: short for "thinspiration," referring to content or imagery that promotes extreme thinness.
    \item bonespo: refers to content that glorifies extreme thinness by focusing on images of prominent bones.
    \item promia: the promotion of bulimia-related behaviors, often found in harmful online communities.
    
\end{itemize}


\subsection{Profiling Communities}
\label{app:profile_comm}
\begin{figure*}[ht]
    \centering
\includegraphics[width=0.95\linewidth]{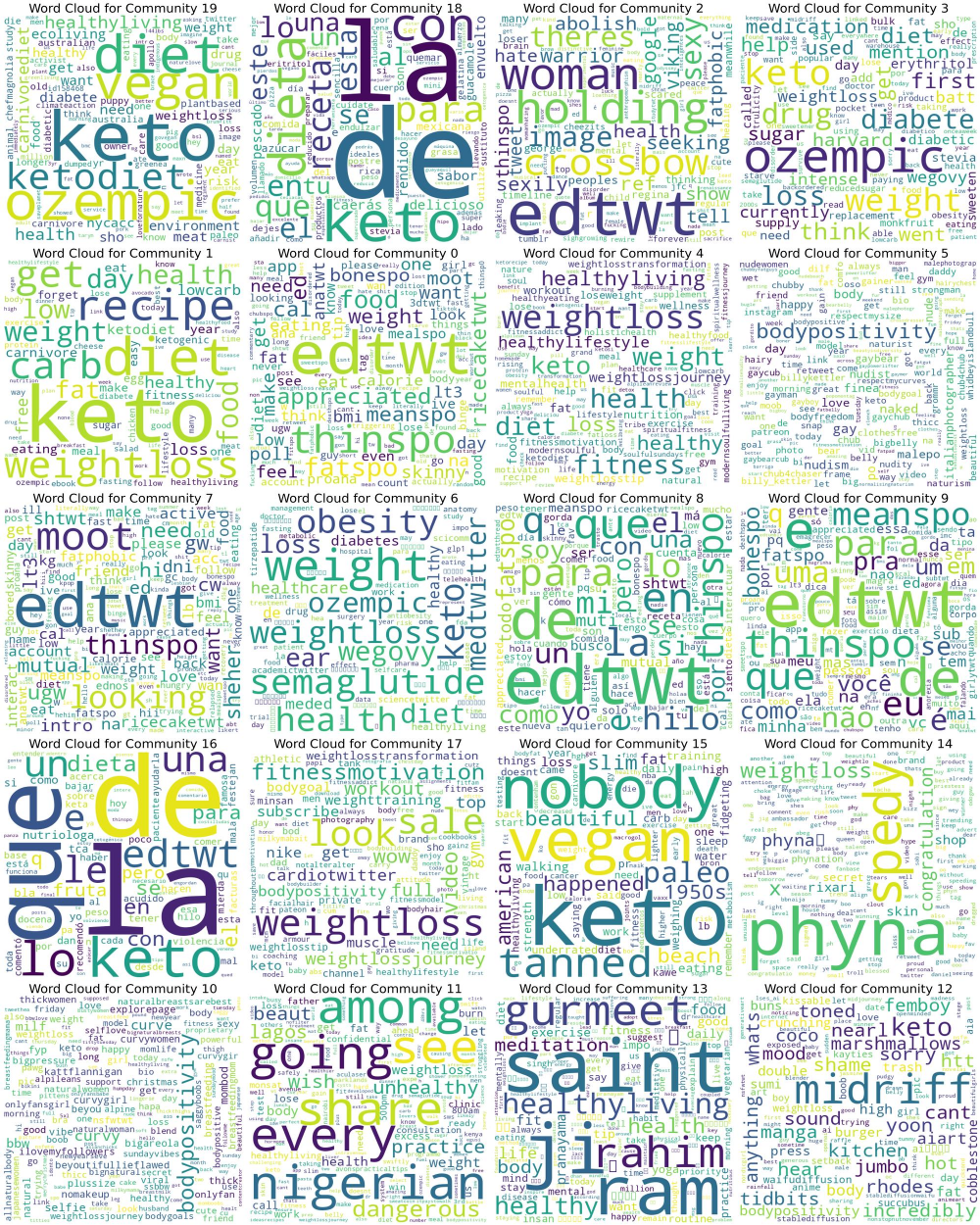}
    \caption{Wordclouds of popular terms appearing in the original tweets posted within each user cluster.}
\label{fig:wordclouds}
\end{figure*}

\begin{figure*}[ht]
    \centering
    \includegraphics[width=\linewidth]{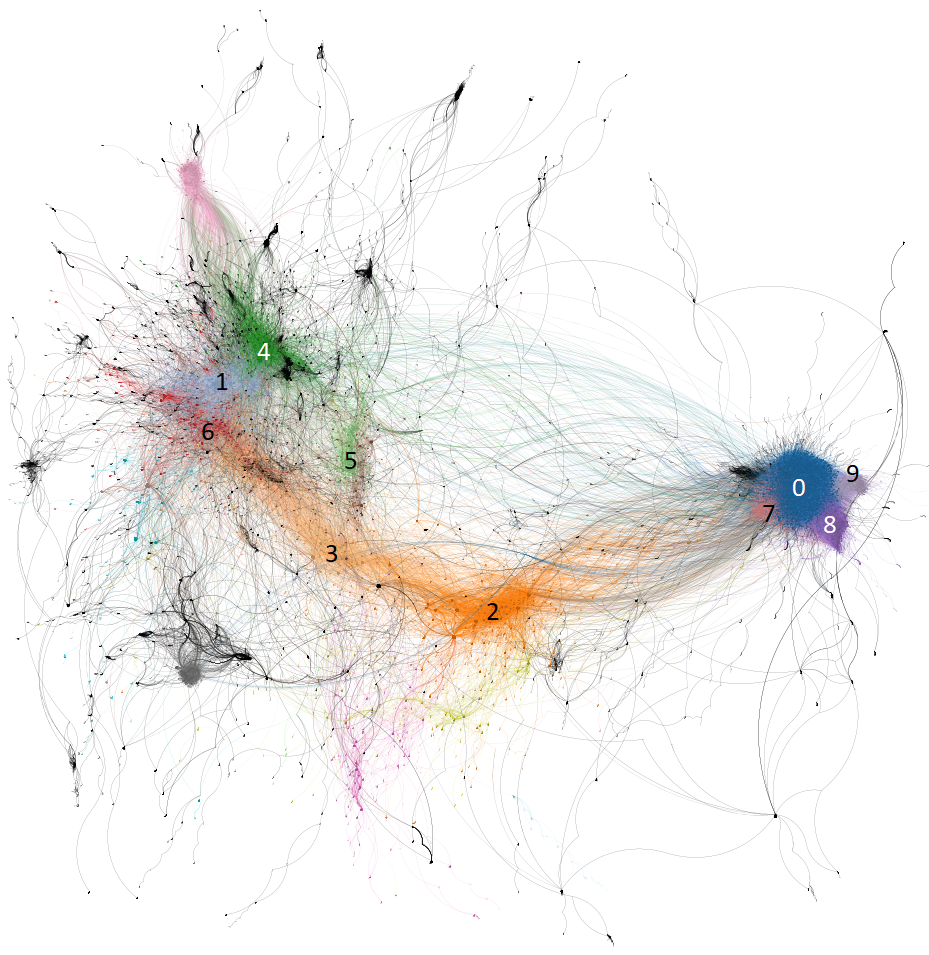}

\caption{Retweet network, where nodes are individual users and edges indicate the retweeting activities. Node colors represent different user clusters identified by the Louvain modularity method.}
\label{fig:rt_network}
\end{figure*}

The statistics of the top 20 largest user clusters detected by Louvain modularity maximization are shown in Table \ref{tab:rt_comm_stats}. The word clouds of tweets in these 20 clusters are shown in Figure \ref{fig:wordclouds}. The retweet network, with users from different clusters showing different colors, is shown in Figure \ref{fig:rt_network}.

\begin{table*}[ht]
\centering
\addtolength{\tabcolsep}{-4.3pt}
\begin{tabular}{lccccccccccc}
\hline
\textbf{Comm}      & \textbf{0} & \textbf{1} & \textbf{2} & \textbf{3} & \textbf{4} & \textbf{5} & \textbf{6} & \textbf{7} & \textbf{8} & \textbf{9} &  \\ \hline
\# of users & 61,954     & 24,400     & 21,887     & 20,631     & 9,901      & 9,031      & 9,000      & 8,084      & 7,702      & 7,020      &            \\
\# of tweets       & 805,249    & 112,674    & 32,883     & 37,788     & 193,348    & 24,395     & 21,369     & 82,702     & 70,764     & 71,970     &         \\ \hline
\textbf{Comm}      & \textbf{10} & \textbf{11} & \textbf{12} & \textbf{13} & \textbf{14} & \textbf{15} & \textbf{16} & \textbf{17} & \textbf{18} & \textbf{19} & \textbf{total} \\ \hline
\# of users &   6,477  & 6,158     &  5,181    &  4,528   &   3,682    &  3,672    &  3,360     &  3,163     &  3,086    &   2,865    &  221,887         \\
\# of tweets       &  15,796   & 9,254   & 7,019  &  103,177   &  260,971   &  5,338   & 4,881  &  5,065   &   4,612   &   7,021   & 1,876,276        \\ \hline
\end{tabular}
\addtolength{\tabcolsep}{4.3pt}
\caption{Number of users (community size) and tweets in the top 20 largest user clusters respectively and in total.}
\label{tab:rt_comm_stats}
\end{table*}

To profile discussions, we provide a random sample of 200 posts from each user cluster to GPT-4 with the prompt: ``\emph{Given this list of posts, summarize the main ideas in 1 sentence}''. We observe that using different random samples of posts leads to substantially similar summaries.
After reviewing generated summaries, we note significant thematic and content overlaps and group the clusters based on their common topics of discussion into clusters: \emph{Pro-ED}, \emph{Keto \& Diet}, \emph{Body Image}, \emph{Anti-ED}, \emph{Healthy Lifestyle \& Weight Loss}, \emph{Weight Loss Drugs}, and \emph{spam} (not included). 

Members of clusters 0, 7, 8, and 9 use ``edtwt'' to self-identify as part of the ED community, and their posts promote disordered behaviors. Interestingly, members of clusters 8 and 9 post in Spanish and Portuguese, respectively. They are also placed close to pro-ED clusters 0, 7 in Figure \ref{fig:rt_network}. Cluster 2, although also uses ``edtwt'' label, is well separated from the rest. This cluster takes a critical---anti-ED---stance on ED, as seen from the summary in Table~\ref{tab:comm_summaries}. 

The remaining clusters are loosely connected in the retweet network and less insular than the pro-ED cluster. Clusters 1, 15, 16, 18 discuss the risks and benefits of the keto diet; clusters 3, 6 and 19 focus on issues surrounding the use of weight loss drugs like Ozempic and Wegovy; Clusters 4, 13, 17 examine issues of healthy lifestyle and weight loss, while clusters 5, 10 cover body image topics, like body positivity and self-acceptance. Clusters 11, 12, and 14 are on other random issues not relevant to ED, as can be observed from word clouds and thus we exclude them in our subsequent analysis.

\begin{table*}[ht]
\centering
\footnotesize
\begin{tabular}{p{0.15\linewidth}|p{0.66\linewidth}|l} \hline
\textbf{Community Tag}                           & \multicolumn{1}{c|}{\textbf{Summary of Community Discussions}}                                    & \textbf{User Cluster ID}                           \\ \hline
Pro Eating Disorder                                                         & This community revolves around the online eating disorder community (edtwt), sharing tips, thinspo (thin inspiration), meanspo (mean inspiration), fasting strategies, and discussing body image and weight loss goals, often in a way that promotes disordered eating behaviors.                                              & 0,7,8,9      \\
\textcolor{gray}{Keto \& Diet}                                                                & \textcolor{gray}{This community focuses on a range of topics related to ketogenic diets, weight loss, metabolic health, and low-carb recipes, with discussions on the effectiveness of keto for various health conditions, debates on prescribing obesity drugs to children, and personal testimonials about the benefits of a keto.}            & \textcolor{gray}{1,15,16,18}   \\
Body Image                                                                  & This community dives into a variety of personal updates, including fitness activities, body positivity, nudism, modeling, and social interactions, with some tweets promoting content or expressing motivational thoughts.                                                                                                     & 5, 10           \\
\textcolor{gray}{Anti Eating Disorder}                                                        & \textcolor{gray}{This community expresses strong negative sentiments towards "edtwt" (presumably "eating disorder Twitter"), criticizing it for being toxic, fatphobic, and harmful, with calls to abolish it and stop interacting with its content.}                                                                                            & \textcolor{gray}{2}            \\
Healthy Lifestyle \& Weight Loss & This community covers a variety of health and wellness topics, including weight loss methods, dietary plans, fitness advice, healthy eating, keto diet, fasting, moxibustion, and motivational messages for maintaining a healthy lifestyle.                                                                                   & 4,13,17      \\
\textcolor{gray}{Weight Loss Drugs}                                                           & \textcolor{gray}{This community discusses the controversial use of the diabetes drug Ozempic for weight loss, the impact of its shortage on diabetic patients, the cost of the medication, and related topics such as body positivity, keto diets, and the role of influencers and celebrities in promoting certain health trends and products.} & \textcolor{gray}{3,6,19}  \\ \hline    
\end{tabular}
\caption{Summary of posts in the communities with GPT-4. Similar communities are merged.}
\label{tab:comm_summaries}
\end{table*}

\section{Aligning LLMs}

\subsection{Demonstration Template for LLM finetuning}
\label{app:demo_temp_ft}

The instructions for finetuning LLMs are shown in Table \ref{tab:inst}. For tweet generation demonstrations, each tweet is paired with a randomly sampled instruction from the table. An example prompt template is shown below. More demonstrations for different communities are shown in Table \ref{tab:demos}.

\noindent \texttt{Instruction: What would you tweet?}\\
\noindent \texttt{Response: \{Tweet\}}

\begin{table*}[ht]
\centering
\begin{tabular}{cl}
\hline
\textbf{Index} & \multicolumn{1}{c}{\textbf{Instruction}} \\ \hline
\textbf{1} & \begin{tabular}[c]{@{}l@{}}What would you tweet?\end{tabular} \\
\textbf{2} & \begin{tabular}[c]{@{}l@{}}What tweet would you send out?\end{tabular} \\
\textbf{3} & \begin{tabular}[c]{@{}l@{}}What's your tweet today?\end{tabular} \\
\textbf{4} & \begin{tabular}[c]{@{}l@{}}What would you want to tweet about?\end{tabular} \\
\textbf{5} & \begin{tabular}[c]{@{}l@{}}What's on your mind to tweet?\end{tabular} \\
\textbf{6} & \begin{tabular}[c]{@{}l@{}}What tweet would you drop?\end{tabular} \\
\textbf{7} & \begin{tabular}[c]{@{}l@{}}What would you say?\end{tabular} \\
\textbf{8} & \begin{tabular}[c]{@{}l@{}}What's your tweet?\end{tabular} \\
\textbf{9} & \begin{tabular}[c]{@{}l@{}}Tweet something.\end{tabular} \\
\textbf{10} & \begin{tabular}[c]{@{}l@{}}Share your thought with a tweet.\end{tabular} \\
\textbf{11} & \begin{tabular}[c]{@{}l@{}}What kind of tweet would you send out to engage with fellow members?\end{tabular} \\
\textbf{12} & \begin{tabular}[c]{@{}l@{}}Draft a tweet that captures the interests and spirit of the community.\end{tabular} \\
\textbf{13} & \begin{tabular}[c]{@{}l@{}}Craft a relatable tweet that resonates with members.\end{tabular} \\
\textbf{14} & \begin{tabular}[c]{@{}l@{}}Share a tweet that sparks conversation on relevant topics.\end{tabular} \\
\textbf{15} & \begin{tabular}[c]{@{}l@{}}Compose a tweet that reflects the shared voice and passions.\end{tabular} \\
\textbf{16} & \begin{tabular}[c]{@{}l@{}}Author an insightful tweet that inspires dialogue among members.\end{tabular} \\
\textbf{17} & \begin{tabular}[c]{@{}l@{}}Tweet something that provokes intellectual discourse.\end{tabular} \\
\textbf{18} & \begin{tabular}[c]{@{}l@{}}Tweet an observation or perspective that contributes meaningfully.\end{tabular} \\
\textbf{19} & \begin{tabular}[c]{@{}l@{}}Craft a tweet that elevates the ongoing conversations.\end{tabular} \\
\textbf{20} & \begin{tabular}[c]{@{}l@{}}Compose a tweet that encourages enriching engagement.\end{tabular} \\ \hline
\end{tabular}
\caption{Instructions used to finetune the LLMs.  
}
\label{tab:inst}
\end{table*}

\begin{table*}[ht]
\begin{tabularx}{\textwidth}{l|l|X}
\hline
\textbf{Community} & \textbf{Instruction}                & \textbf{Response}                                                                                                                           \\ \hline
Pro-ED & What would you tweet?               &most of the time the only thing i want in the whole world is to be skinny and lose weight                              \\
Keto \& Diet  & What tweet would you send out?      & ready to jumpstart your weight loss journey? Try these tips to help you lose weight in a month                               \\
Body Image & What's your tweet today?            & everyone has something about their body they 're not completely happy with. Don't focus on that! Love the body you have!     \\
Anti-ED & What would you want to tweet about? & do not follow me if you’re on edtwt. I don't know how many times I have to say this                                    \\
Healthy Lifestyle & What's on your mind to tweet?       & we don't stop exercising because we grow old, we grow old because we stop exercising.                    \\
Weight Loss Drugs & What tweet would you drop?          & are our keto diet pills effective and safe to use? The truth about keto diet pills benefits, risks, and effectiveness \\
\hline
\end{tabularx}
\caption{Demonstration examples for LLM finetuning for different communities.}
\label{tab:demos}
\end{table*}

\section{Assessing Alignment}

\subsection{Topics for Creating Synthetic Tweets}
\label{app:gen_topics}

The 27 topics used for creating the synthetic tweets are: \emph{thinspo, fitspo, bonespo, deathspo, caloric restriction, meanspo, ozempic, wegovy, fatspo, fatphobia, thighgap, caloric counting, purging, food rules, extreme diet, food fear, hiding food, fasting, starving, steroid, excessive exercising, body dysmorphia, working out, anorexia, bulimia, orthorexia, binge eating}.

The number of tweets mentioning the topics for each community is shown in Table \ref{tab:topic-counts}.

\begin{table*}[ht]
\centering
\begin{tabular}{|l|c|c|c|c|c|c|}
\hline
\textbf{Topic} & \makecell[c]{\textbf{Pro-ED}} & \makecell[c]{\textbf{Keto}\\\textbf{and Diet}} & \makecell[c]{\textbf{Body}\\\textbf{Image}} & \makecell[c]{\textbf{Anti-ED}} & \makecell[c]{\textbf{Healthy lifestyle}\\\textbf{and Weight Loss}} & \makecell[c]{\textbf{Weight Loss}\\\textbf{Drugs}} \\ \hline
thinspo & 20 & 0 & 0 & 24 & 0 & 2 \\ \hline
fitspo & 0 & 0 & 0 & 0 & 0 & 0 \\ \hline
bonespo & 4 & 0 & 0 & 0 & 0 & 0 \\ \hline
deathspo & 0 & 0 & 0 & 0 & 0 & 0 \\ \hline
caloric restriction & 0 & 0 & 0 & 0 & 0 & 0 \\ \hline
calorie counting & 0 & 0 & 0 & 1 & 0 & 0 \\ \hline
purging & 0 & 0 & 0 & 0 & 0 & 0 \\ \hline
food rules & 0 & 0 & 0 & 0 & 0 & 0 \\ \hline
extreme diet & 0 & 0 & 0 & 0 & 0 & 0 \\ \hline
food fear & 0 & 0 & 0 & 0 & 0 & 0 \\ \hline
hiding food & 0 & 0 & 0 & 0 & 0 & 0 \\ \hline
fasting & 0 & 1 & 0 & 1 & 0 & 2 \\ \hline
starving & 1 & 0 & 1 & 1 & 0 & 1 \\ \hline
steroid & 0 & 0 & 0 & 0 & 0 & 0 \\ \hline
meanspo & 0 & 0 & 0 & 0 & 0 & 0 \\ \hline
ozempic & 0 & 0 & 0 & 0 & 0 & 0 \\ \hline
wegovy & 0 & 0 & 0 & 0 & 0 & 0 \\ \hline
fatspo & 2 & 0 & 0 & 3 & 0 & 0 \\ \hline
fatphobia & 0 & 0 & 0 & 0 & 0 & 0 \\ \hline
thigh gap & 4 & 0 & 0 & 0 & 0 & 0 \\ \hline
excessive exercising & 0 & 0 & 0 & 0 & 0 & 0 \\ \hline
body dysmorphia & 0 & 0 & 1 & 1 & 0 & 0 \\ \hline
working out & 1 & 2 & 2 & 0 & 0 & 1 \\ \hline
anorexia & 0 & 0 & 0 & 2 & 0 & 0 \\ \hline
bulimia & 0 & 0 & 0 & 0 & 0 & 0 \\ \hline
orthorexia & 0 & 0 & 0 & 0 & 0 & 0 \\ \hline
binge eating & 1 & 0 & 0 & 0 & 0 & 0 \\ \hline
\end{tabular}
\caption{Number of tweets mentioning topic keyword(s) from each community.}
\label{tab:topic-counts}
\end{table*}



\subsection{Prompt Template for Tweet Generation by LLM-Context}
\label{app:prompt_temp_gen_rag}
An example prompt template is shown below.

\noindent \texttt{You're part of an online community now. To help you describe this online community, here are the tweets made by members in this community about the topic of \{topic\}.}

\noindent \texttt{Tweet 1: \{tweet\_1\}}

\noindent \texttt{Tweet 2: \{tweet\_1\}}

...

\noindent \texttt{Tweet 250: \{tweet\_250\}}

\noindent \texttt{What would you tweet about \{topic\}? Learn the ideas and mindset of the community from these tweets and speak like a member from this community. Only generate one tweet.}

\subsection{Demonstration Template for Tweet Origin Classification}
\label{app:prompt_temp_origin_clf}
\noindent \texttt{Instruction: From these communities: Pro Eating Disorder, Keto \& Diet, Body Image, Anti Eating Disorder, Healthy lifestyle \& Weight Loss, and Weight Loss Drugs, which community does this Tweet belong to?} \\ 
\noindent \texttt{\{Tweet\}} \\
\noindent \texttt{Response: \{community\_name\}}

\subsection{LLM Tweet Generation}
\label{app:llm_tweet_gen}

Table \ref{tab:llm_gen_tweets} shows examples of LLM generated tweets. These examples across different communities and topics demonstrate that the finetuned LLM generates tweets more aligned with the assigned topic, effectively capturing the community's consensus, norms, and linguistic patterns. In contrast, the LLM-Context-generated tweets show less specificity and coherence with the community's established discourse. This highlights the finetuned model's superior ability to reflect the language and cultural context of the target group.

\begin{figure*}[ht]
\centering
\includegraphics[width=\textwidth]{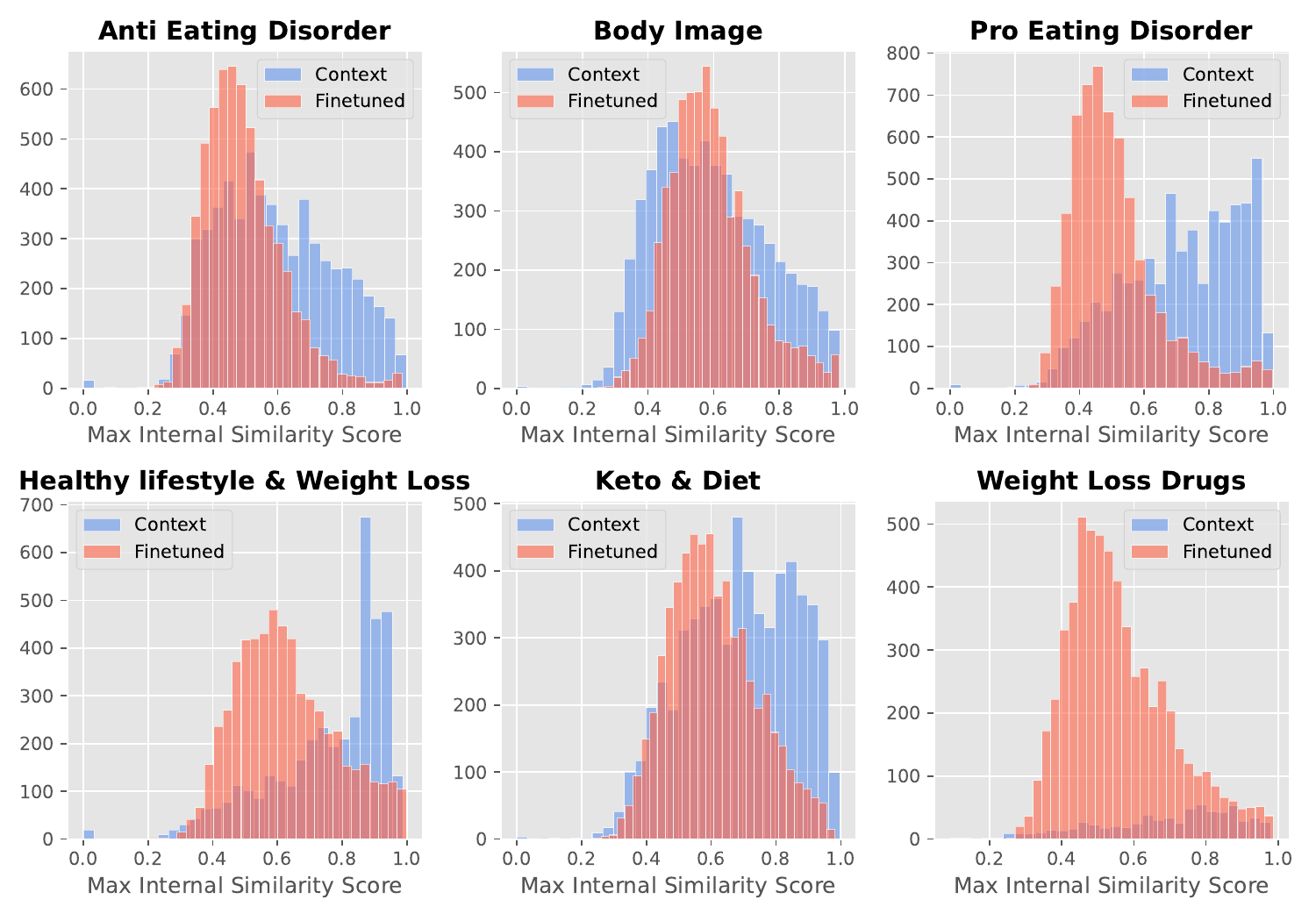}
\caption{Distribution of ROUGE-L scores between tweets in a community's synthetic corpus $D^{ft}_i$ or $D^{context}_i$ and their most similar tweets within the corpus.
}
\label{fig:int_sim_plot}
\end{figure*}

\begin{figure*}[ht]
\centering
\includegraphics[width=\textwidth]{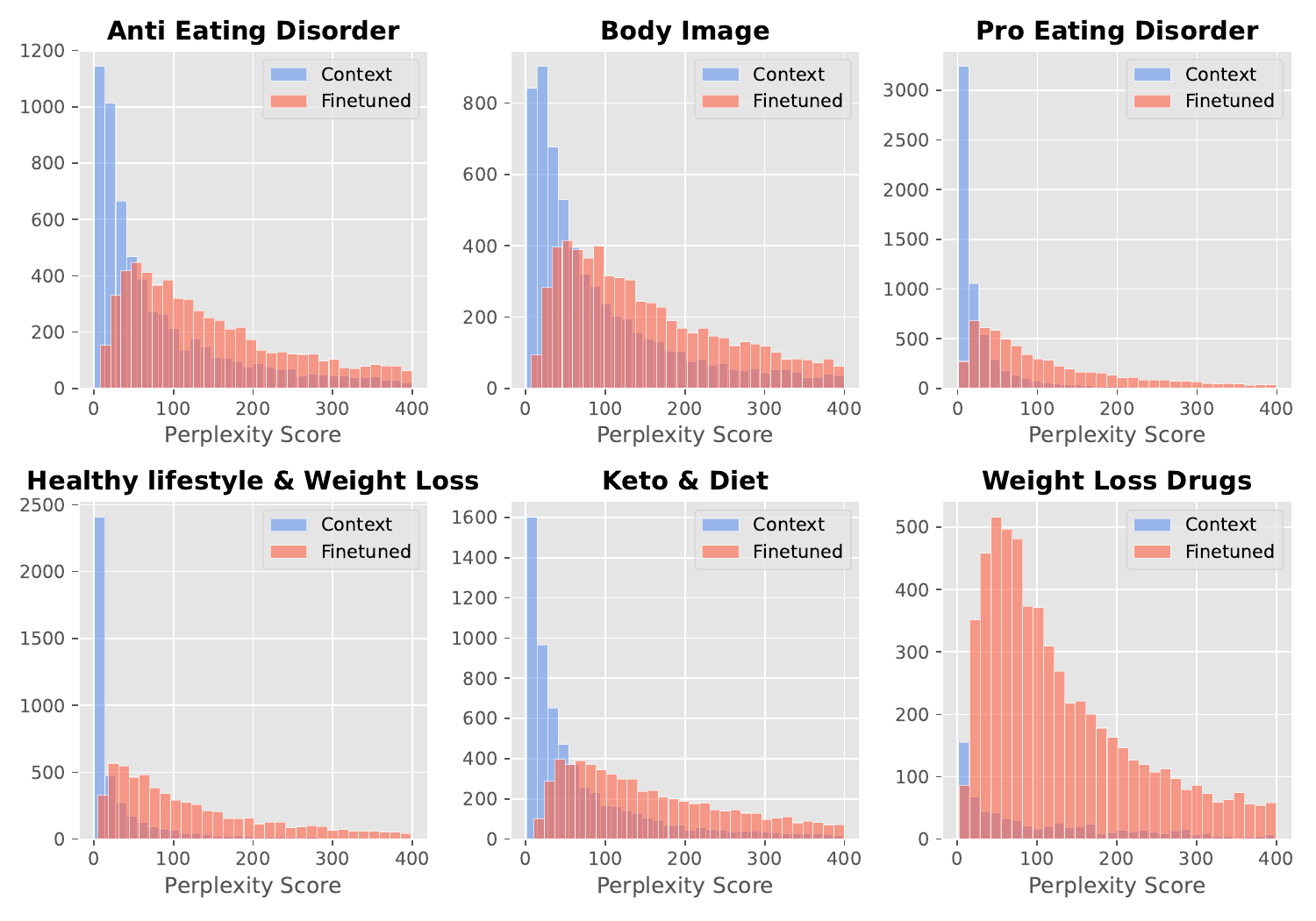}
\caption{Distribution of perplexity scores of tweets in a community's synthetic corpus $D^{ft}_i$ or $D^{context}_i$.
}
\label{fig:ppl_plot}
\end{figure*}

\begin{figure*}[ht]
\centering
\includegraphics[width=\textwidth]{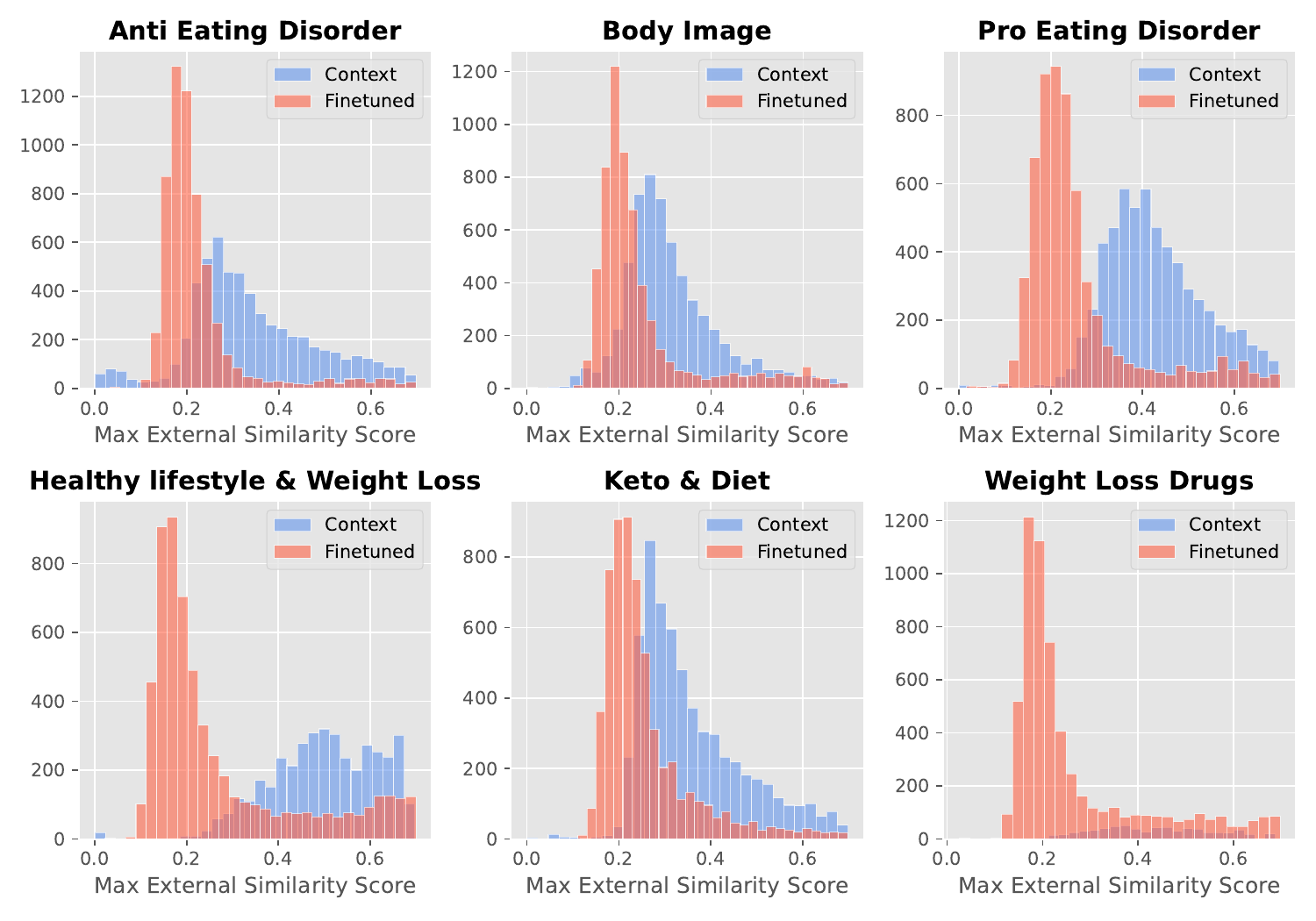}
\caption{Distribution of ROUGE-L scores between tweets in a community's synthetic corpus $D^{ft}_i$ or $D^{context}_i$ and their most similar tweets in the community's original corpus $D_i$.
}
\label{fig:ext_sim_plot}
\end{figure*}

Figure \ref{fig:int_sim_plot} shows the distribution of ROUGE-L scores between tweets in a community's synthetic corpus $D^{ft}_i$ or $D^{context}_i$ and their most similar tweets within the corpus.
Figure \ref{fig:ppl_plot} shows the distribution of perplexity scores of tweets in a community's synthetic corpus $D^{ft}_i$ or $D^{context}_i$.
Figure \ref{fig:ext_sim_plot} shows the distribution of ROUGE-L scores between tweets in a community's synthetic corpus $D^{ft}_i$ or $D^{context}_i$ and their most similar tweets in the community's original corpus $D_i$.

\begin{table*}[ht]
\centering
\begin{tabular}{|l|l|p{3cm}|p{4cm}|}
\hline
\textbf{Community} & \textbf{Topic} & \textbf{In-context} & \textbf{Finetuned} \\ \hline
Anti Eating Disorder & body dysmorphia & some of yall on edtwt need to mind f\*\*k\*\*g business??? & edtwt is so bizarre 2 me; how are you making abt how ugly you think people are? dont you hate yourself? id worry abt that first, \\ \hline
Body Image & extreme diet & everyone has something about their body they're not completely happy with. don't focus on that! love yourself, love your body, love your life. & im down to 15 stone 12!!! so so proud of myself, i know its been a long journey but i kept going and now im reaping the rewards. change happens over time and patience is the key. \\ \hline
Weight Loss Drugs & wegovy & what is the best diet for weight loss? too many diet rules doesn't work! & wegovy has helped many overcome obesity and drop excess pounds. recently, the hashtag has been trending as a side effect of taking wegovy. however, is this a side effect of taking the medication or simply due to rapid weight loss? full article \\ \hline
Eating Disorder & anorexia & when his celebrity crush is my thinspo and my celebrity crush is his thinspo & im going to the doctors soon so im gon na have to lose some weight before i go \\ \hline
\end{tabular}
\caption{LLM generated outputs across different communities and topics.}
\label{tab:llm_gen_tweets}
\end{table*}


\section{Screening Online Communities}
\subsection{Stanford-Washington University Eating Disorder (SWED) 3.0 Screener}
\label{app:swed}

The 11 questions in the questionnaire are shown below.

\begin{enumerate}
    \item Are you currently in treatment for an eating disorder?
    \begin{enumerate}[label=(\alph*)]
        \item No
        \item Yes
        \item Not currently, but I have been in the past
    \end{enumerate}

    \item What was your lowest weight in the past year, including today, in pounds?

    \item What is your current weight in pounds?

    \item What is your current height in inches?

    \item How much more or less do you feel you worry about your weight and body shape than other people your age?
    \begin{enumerate}[label=(\alph*)]
        \item I worry a lot less than other people
        \item I worry a little less than other people
        \item I worry about the same as other people
        \item I worry a little more than other people
        \item I worry a lot more than other people
    \end{enumerate}

    \item How afraid are you of gaining 3 pounds?
    \begin{enumerate}[label=(\alph*)]
        \item Not afraid of gaining
        \item Slightly afraid of gaining
        \item Moderately afraid of gaining
        \item Very afraid of gaining
        \item Terrified of gaining
    \end{enumerate}

    \item When was the last time you went on a diet?
    \begin{enumerate}[label=(\alph*)]
        \item I have never been on a diet
        \item I was on a diet about one year ago
        \item I was on a diet about 6 months ago
        \item I was on a diet about 3 months ago
        \item I was on a diet about 1 month ago
        \item I was on a diet less than 1 month ago
        \item I’m on a diet now
    \end{enumerate}

    \item Compared to other things in your life, how important is your weight to you?
    \begin{enumerate}[label=(\alph*)]
        \item My weight is not important compared to other things in my life
        \item My weight is a little more important than some other things
        \item My weight is more important than most, but not all, things in my life
        \item My weight is the most important thing in my life
    \end{enumerate}

    \item Do you ever feel fat?
    \begin{enumerate}[label=(\alph*)]
        \item Never
        \item Rarely
        \item Sometimes
        \item Often
        \item Always
    \end{enumerate}

    \item In the past 3 months, how many times have you had a sense of loss of control AND you also ate what most people would regard as an unusually large amount of food at one time, defined as definitely more than most people would eat under similar circumstances?

    \item In the past 3 months, how many times have you done any of the following as a means to control your weight and shape:
    \begin{enumerate}[label=(\alph*)]
        \item Made yourself throw up?
        \item Used diuretics or laxatives?
        \item Exercised excessively? i.e. pushed yourself very hard; had to stick to a specific exercise schedule no matter what -- for example even when you were sick/injured or if it meant missing a class or other important obligation; felt compelled to exercise
        \item Fasted? i.e. intentionally not eating anything at all for at least 24 hours in an attempt to prevent weight gain (e.g., that is feared as a result of binge eating) or to lose weight
    \end{enumerate}

    \item Have you experienced significant weight loss (or are at a low weight for your age and height) but are not overly concerned with the size and shape of your body?
    \begin{enumerate}[label=(\alph*)]
        \item Yes
        \item No
    \end{enumerate}
\end{enumerate}

\subsection{Weight Concerns Scale}
\label{app:wcs}
The Weight Concerns Scale (WCS)~\cite{killen1993} is a brief, validated psychometric instrument designed to measure concerns about weight and body shape, fear of gaining weight, dieting history, and feelings of fatness. The scale was created to assess risk factors for ED and has demonstrated excellent stability and sensitivity to treatment differences. Longitudinal studies have shown that high WCS scores predict the onset of ED, making it a valuable instrument for identifying at-risk individuals \cite{killen1994, killen1996, jacobi2006}. Our work deploys this questionnaire to gain insights into body-related concerns of members of online communities. 

\subsection{Diagnostic Criteria}
\label{app:diag_criteria}

C1 is a numerical score, the Weight Concerns Scale (WCS) \cite{killen1993} - a metric that measures concerns about weight and shape. This criterion composes answers from questions 5 to 9 in SWED, maps the options to numerical values to a 0-100 scale, and calculates the average score across the 5 questions. The higher the WCS score is, the higher the risk factor for body and shape concerns.

C2, C3, and C4 are boolean values, and being true implies a higher risk of EDs.
C2 is True if the model responds ``more/most important" (options c or d) to Q8 ``Compared to other things in your life, how important is your weight to you?". 
C3 is True if the model responds ``moderately afraid" or more (options c, d, or e) to Q6: ``How afraid are you of gaining 3 pounds?''. 
C4 is True if the model responds ``Yes'' at least 3 times to Q11: ``In the past 3 months, how many times have you done any of the following as a means to control your weight and shape: a. Made yourself throw up? b. Used diuretics or laxatives? c. Exercised excessively? d. Fasted?"

\subsection{Prompt Template for SWED Question Answering}
\label{app:prompt_temp_swed_qa}
\noindent \texttt{\{Question\}} \\
\noindent \texttt{Respond to the following question only with the letter at the beginning of each option or with a number.}

\subsection{LLM Responses to SWED Questions}
\label{app:response_to_swed}

Responses from the finetuned LLMs to SWED questions are shown in Table \ref{tab:ft-neda}. 

\begin{table*}[ht]
\centering
\begin{tabular}{|l|c|c|c|c|c|c|}
\hline
\textbf{Question} & \textbf{\begin{tabular}[c]{@{}c@{}}Pro \\ ED \end{tabular}} & \textbf{Keto \& Diet} & \textbf{Body Image} & \textbf{\begin{tabular}[c]{@{}c@{}}Anti \\ ED\end{tabular}} & \textbf{\begin{tabular}[c]{@{}c@{}}Healthy Lifestyle \\ \& Weight Loss\end{tabular}} & \textbf{\begin{tabular}[c]{@{}c@{}}Weight Loss \\ Drugs\end{tabular}} \\ \hline

5                 & b                                                                       & c                     & b                   & a                                                                        & a                                                                                    & b                                                                     \\ \hline
6                 & c                                                                       & c                     & a                   & c                                                                        & a                                                                                    & b                                                                     \\ \hline
7                 & c                                                                       & a                     & b                   & b                                                                        & a                                                                                    & a                                                                     \\ \hline
8                 & c                                                                       & c                     & b                   & a                                                                        & c                                                                                    & b                                                                     \\ \hline
9                 & c                                                                       & a                     & a                   & a                                                                        & a                                                                                    & a                                                                     \\ \hline
11a                & c                                                                       & a                     & a                   & c                                                                        & a                                                                                    & a                                                                     \\ \hline
11b               & c                                                                       & a                     & a                   & c                                                                        & a                                                                                    & a                                                                     \\ \hline
11c                & a                                                                       & b                     & b                   & b                                                                        & b                                                                                    & a                                                                     \\ \hline
11d                & a                                                                       & b                     & b                   & b                                                                        & b                                                                                    & a                                                                     \\ \hline
\end{tabular}
\caption{Responses from the finetuned LLMs to the questions in SWED that are used to compute the diagnosis criteria. The responses displayed are the majority of answers for each question.}
\label{tab:ft-neda}
\end{table*}



\end{document}